\documentclass[runningheads]{llncs}
\usepackage{graphicx}
\usepackage{comment}
\usepackage{amsmath,amssymb} 
\usepackage{color}
\usepackage{subfigure}
\usepackage{ulem}
\usepackage[table]{xcolor}
\usepackage{array}
\usepackage{multirow}

\usepackage[pagebackref=true,breaklinks=true,letterpaper=true,colorlinks,bookmarks=false]{hyperref}

 \usepackage[width=122mm,left=12mm,paperwidth=146mm,height=193mm,top=12mm,paperheight=217mm]{geometry}

\begin{document}
\pagestyle{headings}
\mainmatter
\def\ECCVSubNumber{492}  

\title{Corner Proposal Network for Anchor-free, Two-stage Object Detection} 

\titlerunning{Accepted by ECCV 2020}
%
\author{Kaiwen Duan\inst{1}\thanks{\scriptsize This work was done when the first author was interning at Huawei Noah's Ark Lab.} \and
Lingxi Xie\inst{2} \and
Honggang Qi\inst{1} \and
Song Bai\inst{3} \and
Qingming Huang\inst{1,4} \and
Qi Tian\inst{2}}
\authorrunning{Kaiwen Duan et al.}
%

\institute{$^{1}$University of Chinese Academy of Sciences ~~ $^{2}$Huawei Inc.\\
	$^{3}$Huazhong University of Science and Technology ~~ $^{4}$Peng Cheng Laboratory\\
	\email{kaiwenduan@outlook.com}, \email{198808xc@gmail.com},
	\email{\{hgqi,qmhuang\}@ucas.ac.cn}, \email{songbai.site@gmail.com},
	\email{tian.qi1@huawei.com}}

\maketitle
\vspace{-1ex}
\begin{abstract}
The goal of object detection is to determine the class and location of objects in an image. This paper proposes a novel anchor-free, two-stage framework which first extracts a number of object proposals by finding potential corner keypoint combinations and then assigns a class label to each proposal by a standalone classification stage. We demonstrate that these two stages are effective solutions for improving recall and precision, respectively, and they can be integrated into an end-to-end network. Our approach, dubbed Corner Proposal Network (CPN), enjoys the ability to detect objects of various scales and also avoids being confused by a large number of false-positive proposals. On the MS-COCO dataset, CPN achieves an AP of $49.2\%$ which is competitive among state-of-the-art object detection methods. CPN also fits the scenario of computational efficiency, which achieves an AP of $41.6\%$/$39.7\%$ at $26.2$/$43.3$ FPS, surpassing most competitors with the same inference speed. Code is available at \url{https://github.com/Duankaiwen/CPNDet}.

\end{abstract}

\section{Introduction}
\label{Introduction}

Powered by the rapid development of deep learning~\cite{lecun2015deep}, in particular deep convolutional neural networks~\cite{krizhevsky2012imagenet,simonyan2014very,he2016deep}, researchers have designed effective algorithms for object detection~\cite{girshick2014rich}. This is a challenging task since objects can appear in any scale, shape, and position in a natural image, yet the appearance of objects of the same class can be very different.

The two keys to a detection approach are to find objects with different kinds of geometry (\textit{i.e.}, high recall) as well as to assign an accurate label to each detected object (\textit{i.e.}, high precision). Existing object detection approaches are roughly categorized according to how objects are located and how their classes are determined. For the first issue, early research efforts are mostly \textbf{anchor-based}~\cite{girshick2015fast,ren2015faster,cai2018cascade,dai2016r,liu2016ssd,lin2017focal,zhang2018single}, which involved placing a number of size-fixed bounding boxes on the image plane, while this methodology was later challenged by \textbf{anchor-free}~\cite{law2018cornernet,duan2019centernet,kong2020foveabox,yang2019reppoints,zhu2019soft,tian2019fcos} methods which suggested depicting each object with one or few keypoints and the geometry. These potential objects are named proposals, and for each of them, the class label is either inherited from a previous output or verified by an individual classifier trained for this purpose. This brings a debate between the so-called \textbf{two-stage} and \textbf{one-stage} approaches, in which people tend to believe that the former works slower but produces higher detection accuracy.

\begin{figure}[t]
\centering 
\includegraphics[width=3.2cm,height=2.1cm]{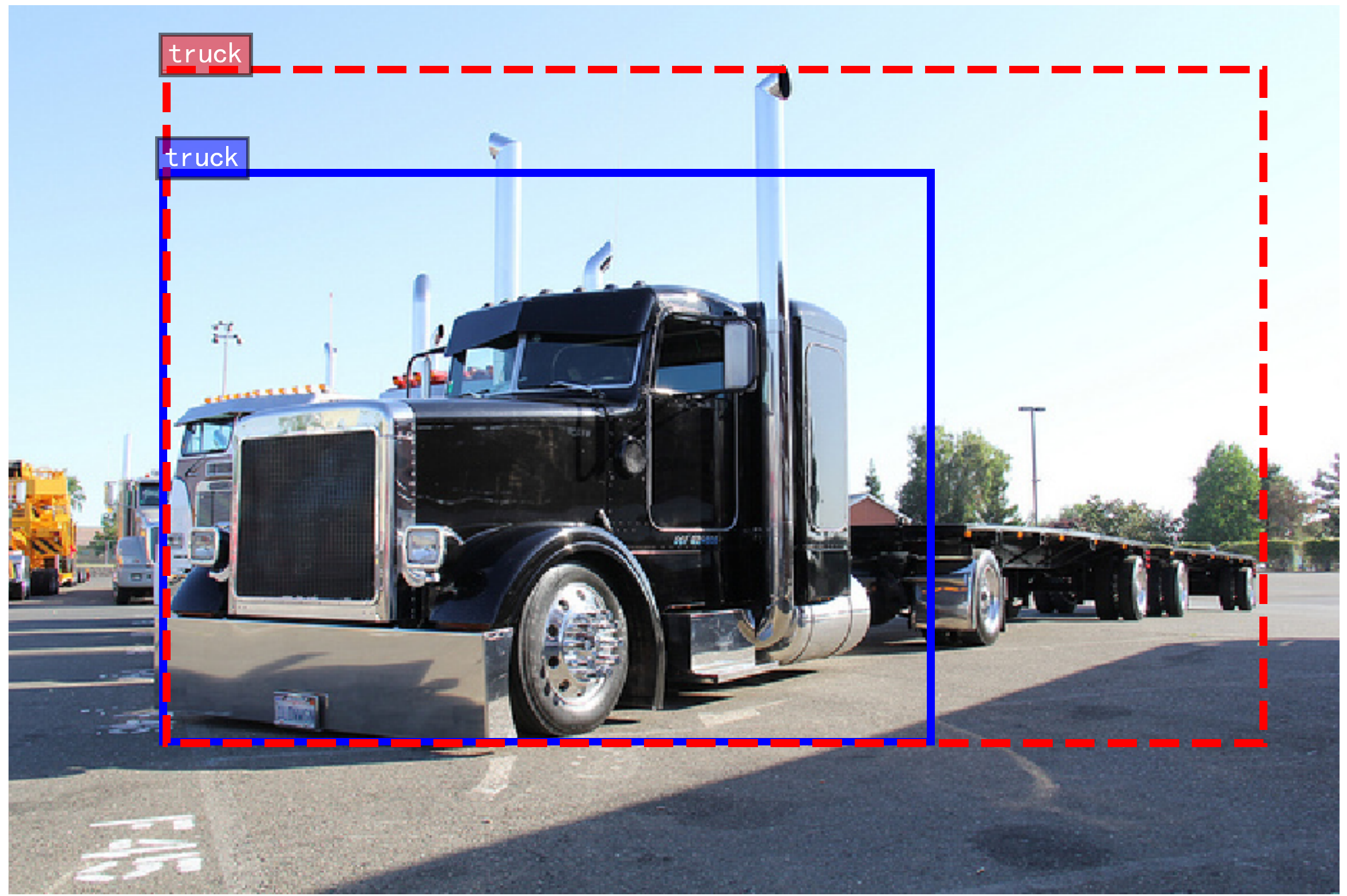}
\includegraphics[width=2.8cm,height=2.15cm]{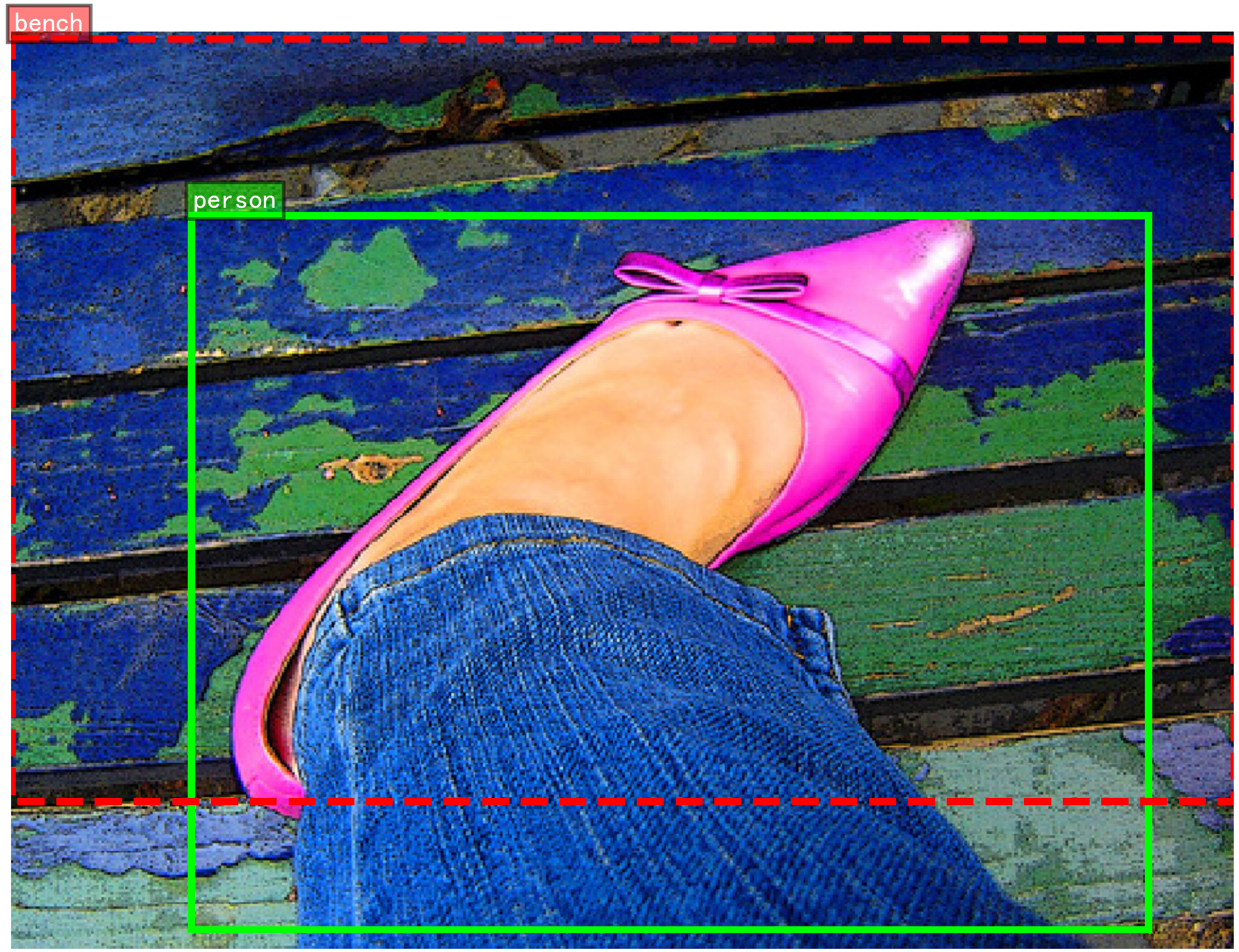}
\includegraphics[width=2.8cm,height=2.1cm]{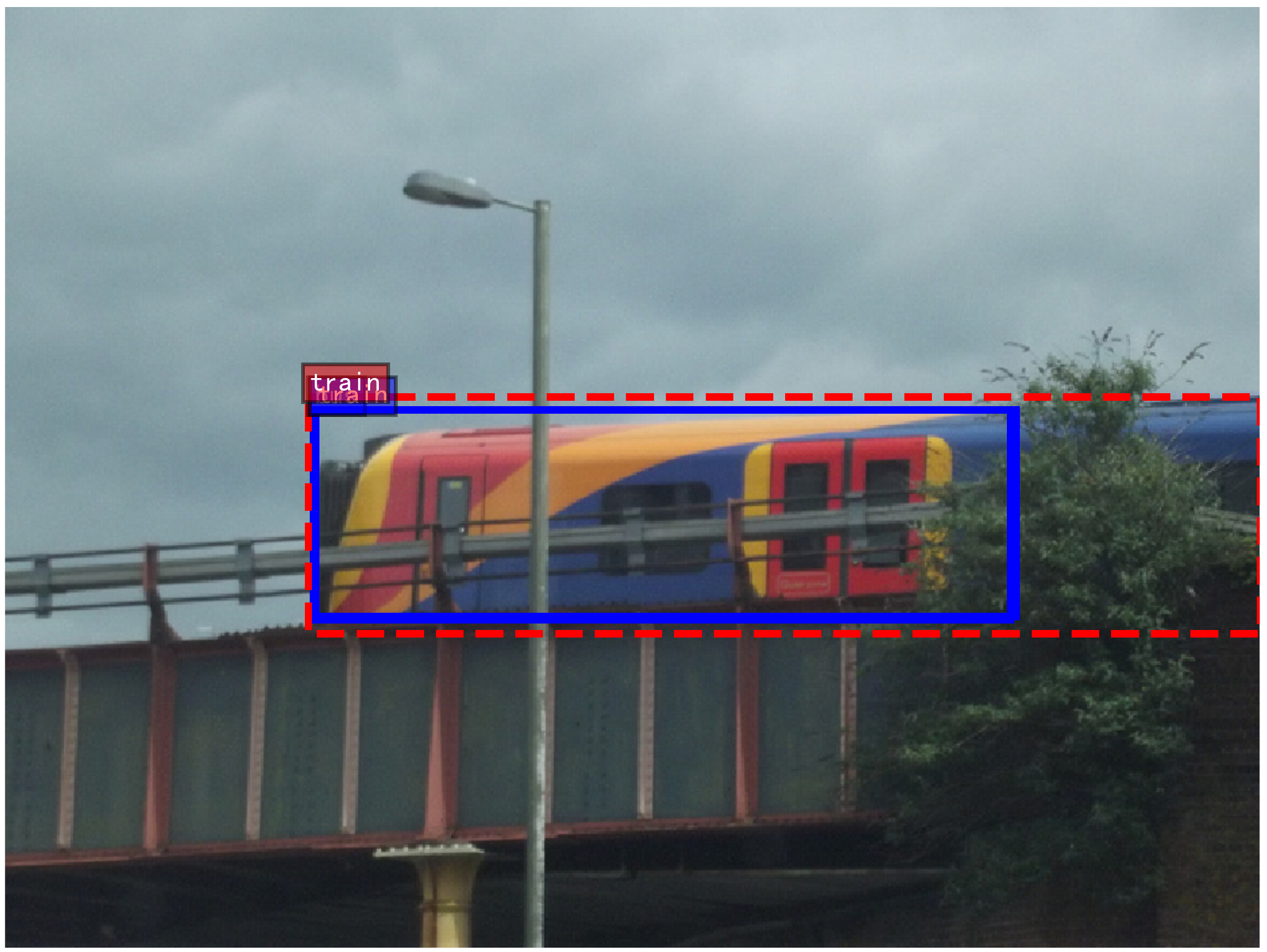}
\includegraphics[width=2.5cm,height=2.1cm]{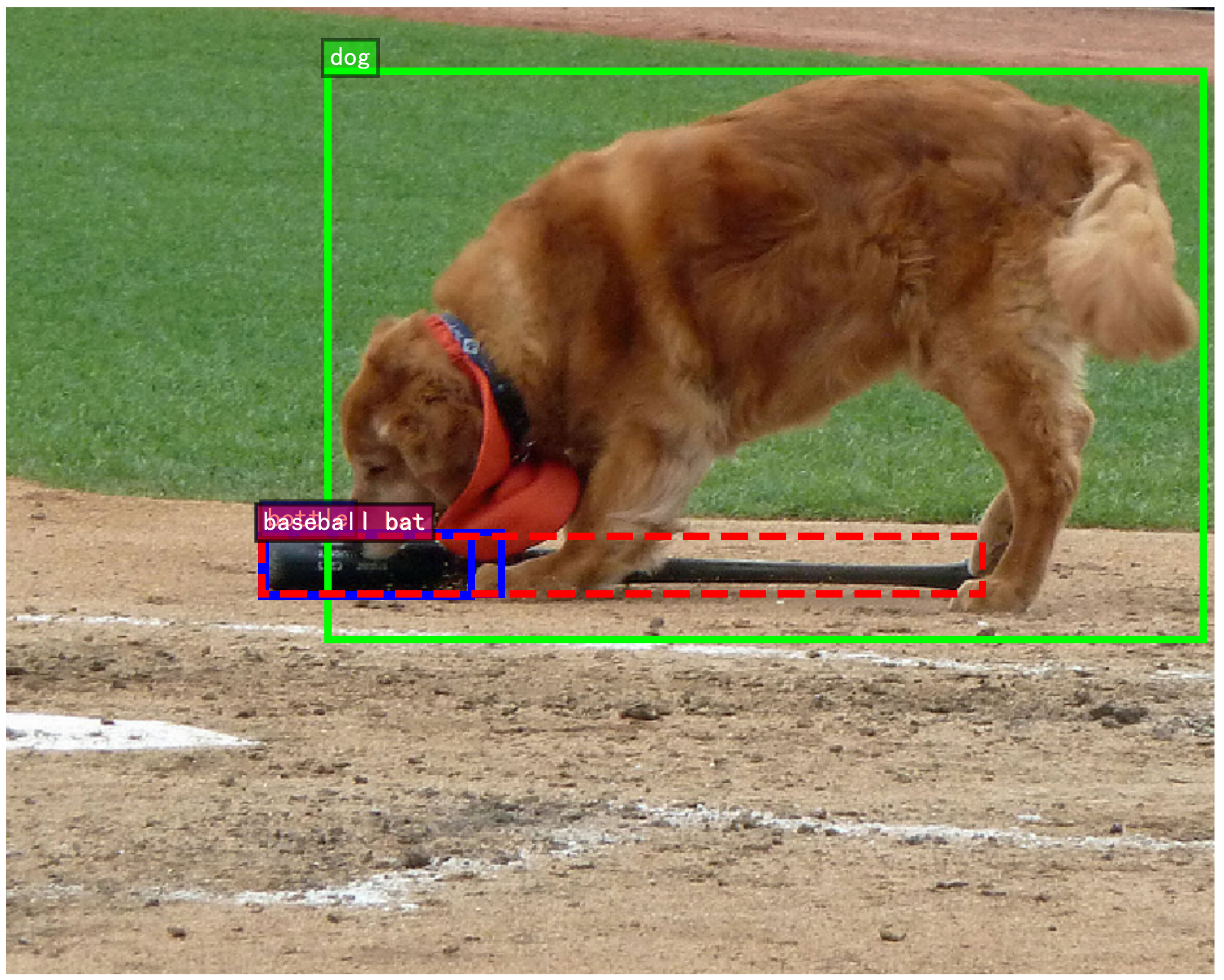}
\includegraphics[width=3.22cm,height=2.05cm]{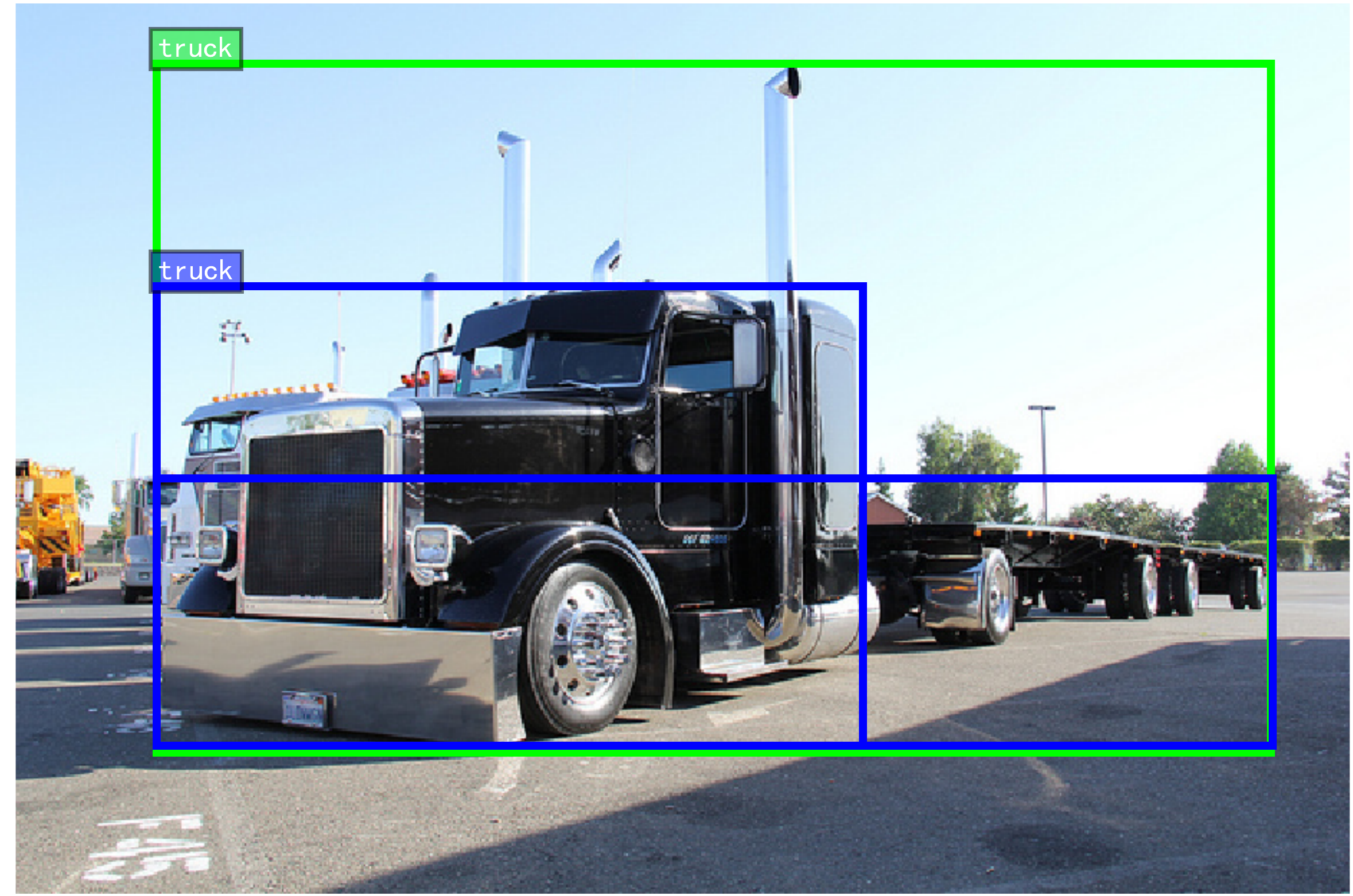}
\includegraphics[width=2.8cm,height=2.1cm]{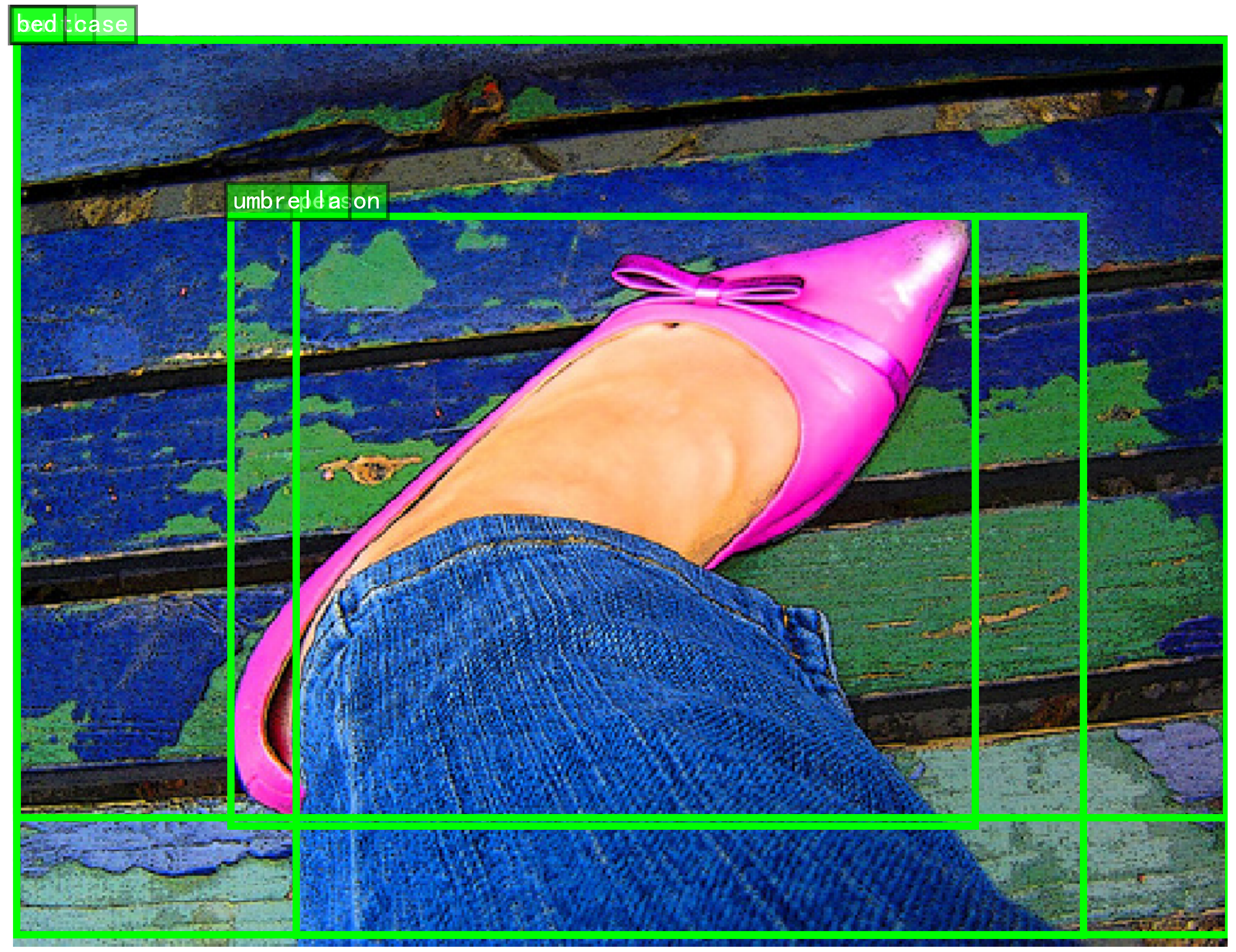}
\includegraphics[width=2.85cm,height=2.1cm]{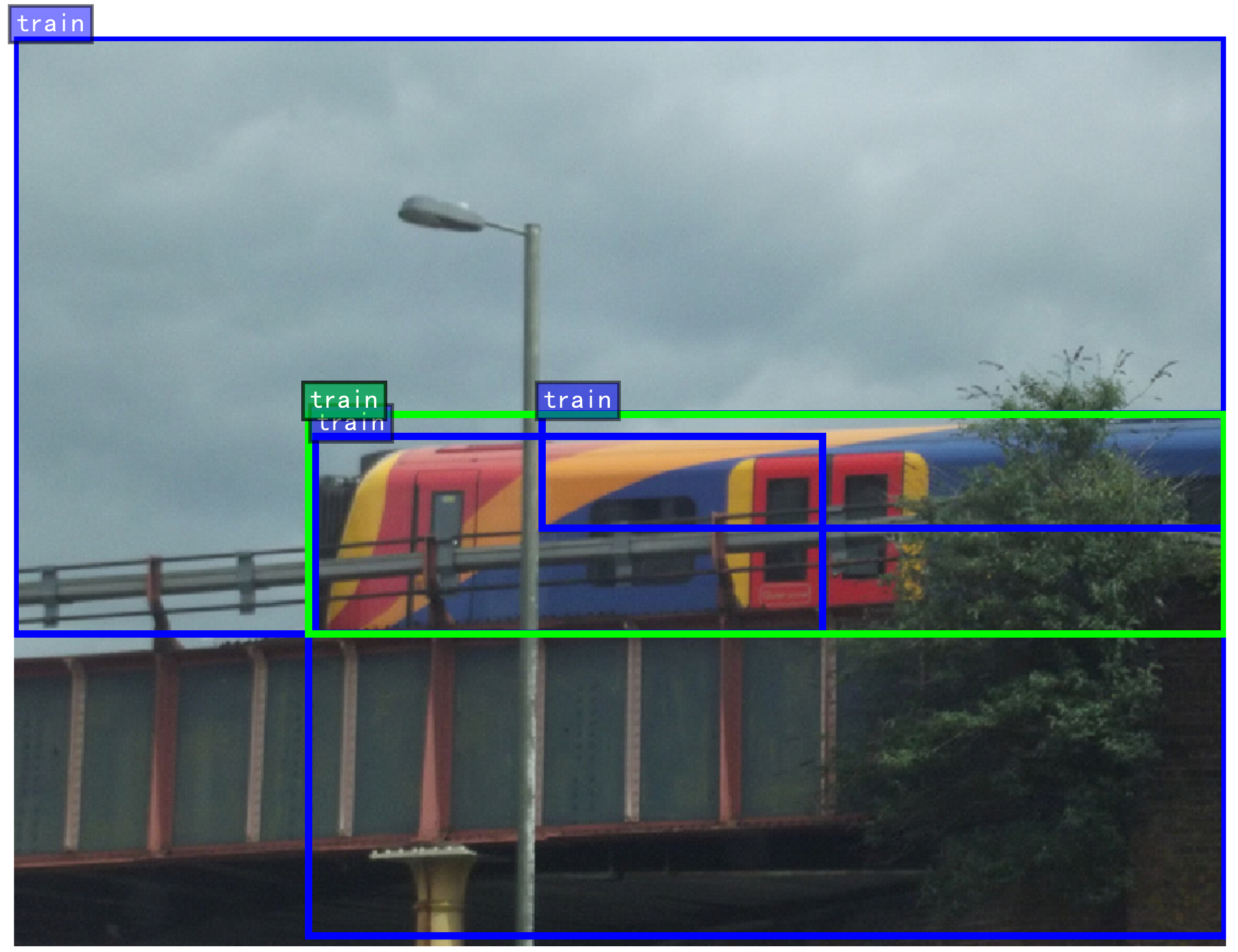}
\includegraphics[width=2.5cm,height=2.05cm]{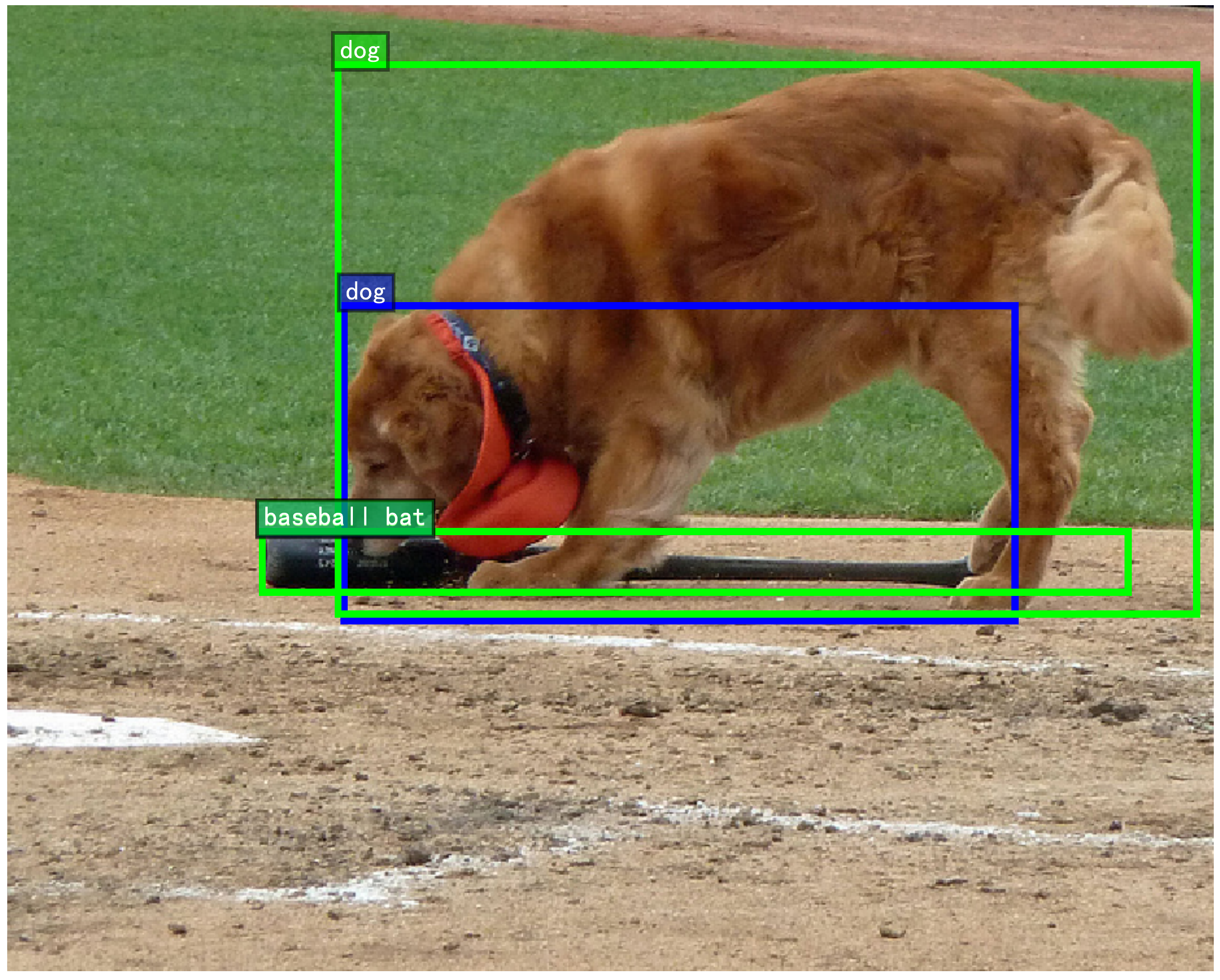}
\caption{Typical errors by existing object detection approaches (best viewed in color). \textbf{Top}: an anchor-based method (\textit{e.g.}, Faster R-CNN~\cite{ren2015faster}) may have difficulty in finding objects with a peculiar shape (\textit{e.g.}, with a very large size or an extreme aspect ratio). \textbf{Bottom}: an anchor-free method (\textit{e.g.}, CornerNet~\cite{law2018cornernet}) may mistakenly group irrelevant keypoints into an object. 
Green, blue and red bounding-boxes indicate true positives, false positives and false negatives, respectively.}
\vspace{-3ex}
\label{fig:detection_errors}
\end{figure}

This paper provides an alternative opinion on the design of object detection approaches. There are two arguments.  \textbf{First}, the recall of a detection approach is determined by its ability to locate objects of different geometries, especially those with rare shapes, and anchor-free methods (in particular, the methods based on locating the border of objects) are potentially better in this task. \textbf{Second}, anchor-free methods often incur a large number of false positives, and thus an individual classifier is strongly required to improve the precision of detection, see Fig.~\ref{fig:detection_errors}. Therefore, we inherit the merits of both anchor-free and two-stage object detectors and design an efficient, end-to-end implementation.

Our approach is named \textbf{Corner Proposal Network} (CPN). It detects an object by locating the top-left and bottom-right corners of it and then assigning a class label to it. We make use of the keypoint detection method of CornerNet~\cite{law2018cornernet} but, instead of grouping the keypoints with keypoint feature embedding, enumerate all valid corner combinations as potential objects. This leads to a large number of proposals, most of which are false positives. We then train a classifier to discriminate real objects from incorrectly paired keypoints based on the corresponding regional features. There are two steps for classification, with the first one, a binary classifier,  filtering out a large part of proposals (\textit{i.e.}, that do not correspond to objects), and the second one, with stronger features, re-ranking the survived objects with multi-class classification scores.

The effectiveness of CPN is verified on the MS-COCO dataset~\cite{lin2014microsoft}, one of the most challenging object detection benchmarks. Using a $104$-layer stacked Hourglass network~\cite{newell2016stacked} as the backbone, CPN reports an AP of $49.2\%$, which outperforms the previously best anchor-free detectors, CenterNet~\cite{duan2019centernet}, by a significant margin of $2.2\%$. In particular, CPN enjoys even larger accuracy gain in detecting objects with peculiar shapes (\textit{e.g.}, very large or small areas or extreme aspect ratios), demonstrating the advantage of using anchor-free methods for proposal extraction. Last but not least, CPN can also fit scenarios that desire for network efficiency. Working on a lighter backbone DLA-34~\cite{yu2018deep} and switching off image flip in inference, CPN achieves $41.6\%$ at $26.2$ FPS or $39.7\%$ at $43.3$ FPS, surpassing most competitors with the same inference speed.
\vspace{-0.5ex}
\section{Related Work}\label{Related_Work}
\vspace{-0.5ex}

Object detection is an important yet challenging problem in computer vision. It aims to obtain a tight bounding-box as well as a class label for each object in an image. In recent years, with the rapid development of deep learning, most powerful object detection methods are based on training deep neural networks~\cite{girshick2014rich,girshick2015fast}. According to the way of determining the geometry and class of an object, existing detection approaches can be roughly categorized into anchor-based and anchor-free methods.

An \textbf{anchor-based approach} starts with placing a large number of anchors, which are regional proposals with different but fixed scales and shapes, and are uniformly distributed on the image plane. These anchors are then considered as object proposals and an individual classifier is trained to determine the objectness as well as the class of each proposal~\cite{ren2015faster}. Beyond this framework, researchers made efforts in two aspects, namely, improving the basic quality of regional features extracted from the proposal, and arriving at a better alignment between the proposals and features. For the first type of efforts, typical examples include using more powerful network backbones~\cite{he2016deep,sun2019deep,huang2017densely} and using hierarchical features to represent a region~\cite{lin2017feature,ren2015faster,liu2016ssd}. Regarding the second type, there exist methods to align anchors to features~\cite{zhang2018single,zhang2019freeanchor}, align features to anchors~\cite{dai2017deformable,chen2019revisiting}, and adjust the anchors after classification has been done~\cite{ren2015faster,liu2016ssd,cai2018cascade}.

Alternatively, an \textbf{anchor-free approach} does not assume the objects to come from uniformly distributed anchors. Early efforts including DenseBox~\cite{huang2015densebox} and UnitBox~\cite{yu2016unitbox} proved that the detectors can achieve the detection task without anchors. Recently, anchor-free approaches have been greatly promoted by the development of keypoint detection~\cite{newell2017associative} and the assist of the focal loss~\cite{lin2017focal}. The fundamental of anchor-free approaches is usually one or few keypoints. Depending on how keypoints are used for object depiction, anchor-free approaches can be roughly categorized into point-grouping detectors and point-vector detectors. Point-grouping detectors, including CornerNet~\cite{law2018cornernet}, CenterNet~\cite{duan2019centernet}, ExtremeNet~\cite{zhou2019bottom}, \textit{etc,} group more than one keypoints into an object, while the point-vector detectors such as FCOS~\cite{tian2019fcos}, CenterNet~\cite{zhou2019objects}, FoveaBox~\cite{kong2020foveabox}, SAPD~\cite{zhu2019soft}, \textit{etc.}, use a keypoint and a vector of object geometry (\textit{e.g.}, the width, height, or its distance to the borders) to determine the shape of objects.

Based on the object proposals, it remains to determine whether each proposal is an object and what class of object it is. There is also discussion on using \textbf{two-stage} and \textbf{one-stage} detectors for object detection. A two-stage detector~\cite{ren2015faster,he2017mask,cai2018cascade,pang2019libra,li2019scale} refers to an individual classifier is trained for this purpose, while a one-stage detector~\cite{redmon2016you,liu2016ssd,lin2017focal,zhang2018single,chen2019revisiting} mostly uses classification cues from the previous stage. Two-stage detectors are often more accurate but slower, compared to one-stage detectors. To accelerate it, an efficient method is to partition classification into two steps~\cite{dai2016r,lin2017feature,he2017mask,cai2018cascade,pang2019libra}, with the first step filtering out most easy false positives, and the second step using heavier computation to assign each survived proposal a class label.
\section{Our Approach}
\label{Approach}
Object detection starts with an image $\mathbf{I}$ denoted by raw pixels, on which a few rectangles, often referred to as bounding-boxes, that tightly covers the objects are labeled with a class label. Denote a ground-truth bounding-box as $\mathbf{b}_n^\star$, ${n}={1,2,\ldots,N}$, the corresponding class label as $c_n^\star$, and $\mathbf{I}_n^\star$ represents the image region within the corresponding bounding-box. The goal is to locate a few bounding-boxes, $\mathbf{b}_m$, ${m}={1,2,\ldots,M}$, and assign each one with a class label, $c_m$, so that the sets of $\left\{\mathbf{b}_n^\star,c_n^\star\right\}_{n=1}^N$ and $\left\{\mathbf{b}_m,c_m\right\}_{m=1}^M$ are as close as possible.
\vspace{-2ex}
\subsection{Anchor-based or Anchor-free? One-stage or Two-stage?}
\label{Approach:Choices}

We focus on two important choices of object detection, namely, whether to use anchor-based or anchor-free methods for proposal extraction, and whether to use one-stage or two-stage methods for determining the class of proposals. Based on these discussions, we present a novel framework in the next subsection.

We first investigate anchor-based \textit{vs.} anchor-free methods. Anchor-based methods first place a number of anchors on the image as object proposals and then use an individual classifier to judge the objectness and class of each proposal. Most often, each anchor is associated with a specific position on the image and its size is fixed, although the following process named bounding-box regression can slightly change its geometry. Anchor-free methods do not assume the objects to come from anchors of relatively fixed geometry, and instead, locate one or few keypoints of an object and determine its geometry and/or class afterward.

Our core opinion is that \textbf{anchor-free methods have better flexibility of locating objects with arbitrary geometry, and thus a higher recall.} This is mainly due to the design nature of anchors, which is mostly empirical (\textit{e.g.}, to reduce the number of anchors and improve efficiency, only common object sizes and shapes are considered), the detection algorithm is potential of lower flexibility and objects with a peculiar shape can be missing. Typical examples are shown in Fig.~\ref{fig:detection_errors}, and a quantitative study is provided in Table~\ref{tab:false_negatives}. We evaluate four object proposal extraction methods as well as our work on the MS-COCO validation dataset and show that anchor-free methods often have a higher overall recall, which is mainly due to their advantages in two scenarios. \textbf{First}, when the object is very large, \textit{e.g.}, larger than $400^2$ pixels, Faster R-CNN, an anchor-based approach, does not report a much higher recall. This is not expected because large objects should be easier to detect, as the other three anchor-free methods suggest. \textbf{Second}, Faster R-CNN suffers a very low recall when the aspect ratio of the object becomes peculiar, \textit{e.g.}, 
$5:1$ and $8:1$, in which cases the recalls are significantly lower than CornerNet~\cite{law2018cornernet} and CenterNet~\cite{duan2019centernet}, because no pre-defined anchors (also used in other variants~\cite{cai2018cascade,liu2016ssd,pang2019libra}) can fit these objects. A similar 
phenomenon is also observed in FCOS, an anchor-free approach which represents an object by a keypoint and the distance to the border, because it is difficult to predict an accurate distance when the border is far from the center. Since CornerNet and CenterNet group corner (and center) keypoints into an object, they somewhat get rid of this trouble.
Therefore, we choose anchor-free methods, in particular, \textbf{point-grouping} methods (CornerNet and CenterNet), to improve the recall of object detection. Moreover, we report the corresponding results of CPN, the method proposed in this paper, which demonstrates that CPN inherits the merits of CenterNet and CornerNet and has better flexibility of locating objects, especially with peculiar shapes.

\begin{table}[!t]
\begin{center}
\caption{Comparison among the average recall (AR) of anchor-based and anchor-free detection methods. Here, the average recall is recorded for targets of different aspect ratios and different sizes. To explore the limit of the average recall for each method, we exclude the impacts of bounding-box categories and sorts on recall, and compute it by allowing at most $1000$ object proposals. $\mathrm{AR_{1+}}$, $\mathrm{AR_{2+}}$, $\mathrm{AR_{3+}}$ and $\mathrm{AR_{4+}}$ denote box area in the ranges of $\left(96^{2}, 200^{2}\right]$, $\left(200^{2}, 300^{2}\right]$, $\left(300^{2}, 400^{2}\right]$, and $\left(400^{2}, +\infty\right)$, respectively. `X' and `HG' stand for ResNeXt and Hourglass, respectively.}
\label{tab:false_negatives}
\resizebox{1\textwidth}{!}{
\begin{tabular}{|l|c|c|cccc|cccc|}
\hline
Method & Backbone & $\mathrm{AR}$ & $\mathrm{AR_{1+}}$ & $\mathrm{AR_{2+}}$ & $\mathrm{AR_{3+}}$ & $\mathrm{AR_{4+}}$ & $\mathrm{AR_{5:1}}$ & $\mathrm{AR_{6:1}}$ & $\mathrm{AR_{7:1}}$ & $\mathrm{AR_{8:1}}$\\
\hline\hline
Faster R-CNN~\cite{ren2015faster} & X-101-64x4d & 57.6 & 73.8 & 77.5 & 79.2 & 86.2 & 43.8 & 43.0 & 34.3 & 23.2\\
\hline\hline
FCOS~\cite{tian2019fcos} & X-101-64x4d & 64.9 & 82.3 & 87.9 & 89.8 & 95.0 & 45.5 & 40.8 & 34.1 & 23.4\\
\hline
CornerNet~\cite{law2018cornernet} & HG-104 & 66.8 & 85.8 & 92.6 & 95.5 & 98.5 & 50.1 & 48.3 & 40.4 & 36.5\\
CenterNet~\cite{duan2019centernet} & HG-104 & 66.8 & 87.1 & 93.2 & 95.2 & 96.9 & 50.7 & 45.6 & 40.1 & 32.3\\
\hline
CPN (this work) & HG-104 & 68.8 & 88.2 & 93.7 & 95.8 & 99.1 & 54.4 & 50.6 & 46.2 & 35.4\\
\hline
\end{tabular}}
\end{center}
\vspace{-5ex}
\end{table}

\begin{table}[!b]
\begin{center}
\vspace{-3ex}
\caption{Anchor-free detection methods such as CornerNet and CenterNet suffer a large number of false positives and can benefit from incorporating richer semantics for judgment. Here, $\mathrm{AP_{original}}$, $\mathrm{AP_{refined}}$, and $\mathrm{AP_{correct}}$ indicate the AP of the original output, after non-object proposals are removed, and after the correct label is assigned to each survived proposal. Both $\mathrm{AP_{refined}}$ and $\mathrm{AP_{correct}}$ require ground-truth labels.}
\label{tab:false_positives}
\begin{tabular}{|l|c|ccc|ccc|}
\hline
Method & Backbone & $\mathrm{AP_{original}}$ & $\mathrm{AP_{refined}}$ & $\mathrm{AP_{correct}}$ & $\mathrm{AR_{original}^{100}}$ & $\mathrm{AR_{refined}^{100}}$ & $\mathrm{AR_{correct}^{100}}$ \\
\hline\hline
CornerNet~\cite{law2018cornernet}  & HG-52 & 37.6 & 53.8 & 60.3 & 56.7 & 65.3 & 69.2 \\
CornerNet~\cite{law2018cornernet}  & HG-104 & 41.0 & 56.6 & 61.6 & 59.1 & 67.0 & 70.4 \\
\hline
CenterNet~\cite{duan2019centernet} & HG-52 & 41.3 & 51.9 & 56.6 & 59.5 & 61.1 & 64.2 \\
CenterNet~\cite{duan2019centernet} & HG-104 & 44.8 & 55.3 & 59.9 & 62.2 & 65.1 & 68.4 \\
\hline
\end{tabular}
\end{center}
\end{table}

However, anchor-free methods free the constraints of finding object proposals, it encounters a major difficulty of building a close relationship between keypoints and objects, since the latter often requires richer semantic information. As shown in Fig.~\ref{fig:detection_errors}, lacking semantics can incur a large number of false positives and thus harms the precision of detection. We take CornerNet~\cite{law2018cornernet} and CenterNet~\cite{duan2019centernet} with potentially high recalls as examples. As shown in Table~\ref{tab:false_positives}, the CornerNets with $52$-layer and $104$-layer Hourglass networks achieved APs of $37.6\%$ and $41.0\%$ on the MS-COCO validation dataset, while many of the detected `objects' are false positives. Either when we remove the non-object proposals or assign each preserved proposal with a correct label, the detection accuracy goes up significantly. This observation also holds on CenterNet~\cite{duan2019centernet}, which added a center point to filter out false positives but obviously did not remove them all. \textit{To further alleviate this problem, we need to inherit the merits of two-stage methods, which extract the features within proposals and train a classifier to filter out false positives.}


\vspace{-2ex}
\subsection{The Framework of Corner Proposal Network}
\label{Approach:Framework}

\begin{figure}[!t]
\centering 
\includegraphics[width=1\textwidth]{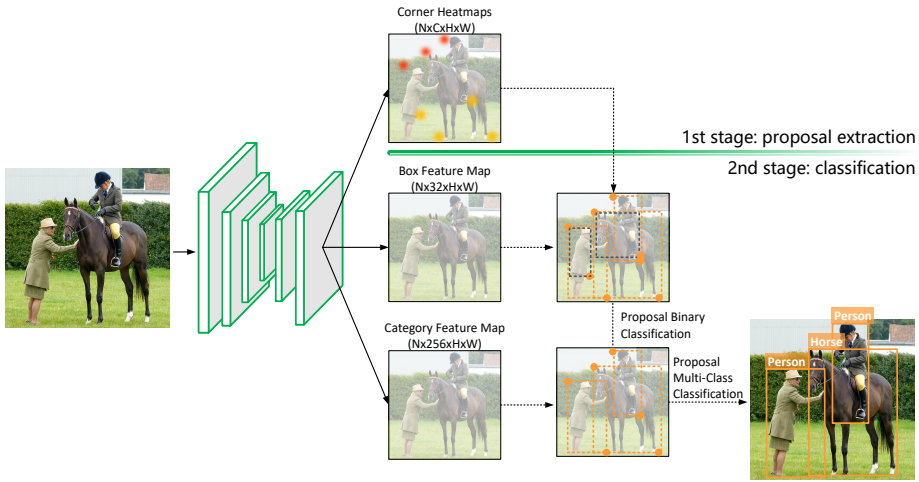}
\vspace{-5ex}
\caption{The overall pipeline of Corner Proposal Network (CPN). It first predicts corners to compose a number of object proposals, and then applies two-step classification to filter out false positives and assign a class label for each survived proposal.}
\label{fig:pipeline}
\vspace{-2ex}
\end{figure}

Motivated by the above analysis, the goal of our approach is to integrate the advantages of anchor-free methods and alleviate their drawbacks by leveraging the mechanism of discrimination from two-stage methods. We present a new framework named \textbf{Corner-Proposal-Network} (CPN). It uses an anchor-free method to extract object proposals followed by efficient regional feature computation and classification to filter out false positives. Fig.~\ref{fig:pipeline} shows the overall pipeline which contains two stages, and details of the two stages are elaborated as follows.

\noindent$\bullet$\quad\textbf{Stage 1: Anchor-free Proposals with Corner Keypoints}

The first stage is an anchor-free proposal extraction process, in which we assume that each object is located by two keypoints determining its top-left and bottom-right corners. We follow CornerNet~\cite{law2018cornernet} to locate an object with a pair of keypoints located in its top-left and bottom-right corners, respectively. For each class, we compute two heatmaps (\textit{i.e.}, the top-left heatmap and the bottom-right heatmap, each value on a heatmap indicates the probability that a corner keypoint occurs in the corresponding position) with a $4\times$-reduced resolution compared to the original image. The heatmaps are equipped with two loss terms, namely, a focal loss $\mathcal{L}_\mathrm{det}^\mathrm{corner}$ to locate the keypoint on the heatmap and a offset loss $\mathcal{L}_\mathrm{offset}^\mathrm{corner}$ to learn its offset to the accurate corner position. After heatmaps are computed, a fixed number of keypoints ($K$ top-left and $K$ bottom-right) are extracted from all heatmaps. Each corner keypoint is equipped with a class label.

Next, each valid pair of keypoints defines an object proposal. Here by valid we mean that two keypoints belong to the same class (\textit{i.e.}, extracted from the top-left heatmap and the bottom-right heatmap of the same class), and the $x$ and $y$ coordinates of the top-left point are smaller than that of the bottom-right point, respectively. This leads to a large number of false positives (incorrectly paired corner keypoints) on each image, and we leave the task of discriminating and classifying these proposals to the second stage.

As a side comment, we emphasize that although we extract object proposals based on CornerNet, the follow-up mechanism of determining objectness and class is quite different. CornerNet generates objects by projecting the keypoints to a one-dimensional space, and grouping keypoints with closely embedded numbers into the same instance. We argue that the embedding process, while necessary under the assumption that no additional computation can be used, can incur significant errors in pairing keypoints. In particular, there is no guarantee that the embedding function (assigning a number to each object) is learnable, and more importantly, the loss function only works in each training image to force the embedded numbers of different objects to be separated, but this mechanism often fails to generalize to unseen scenarios, \textit{e.g.}, even when multiple training images are simply concatenated together, the embedding function that works well on separate images can fail dramatically. Differently, our method determines object instances using an individual classifier, which makes full use of the internal features to improve accuracy. Please refer to Table~\ref{tab:classifier_embedding} for the advantage of an individual classifier over instance embedding.

\noindent$\bullet$\quad\textbf{Stage 2: Two-step Classification for Filtering Proposals}

Thanks to the high resolution of the keypoint heatmap and a flexible mechanism of grouping keypoints, the detected objects can be of an arbitrary shape, and the upper-bound of recall is largely improved. However, this strategy increases the number of proposals and thus brings two problems: a considerable amount of false positives as well as computational costs to filter out them. To solve this issue, an efficient, two-step classification method is designed for the second stage, which first removes $80\%$ of proposals with a light-weighted binary classifier, and then applies a fine-level classifier to determine the class label of each survived proposal.

Let $M$ be the total number of object proposals generated by $K$ top-left and $K$ bottom-right keypoints. We follow CenterNet~\cite{duan2019centernet} to set $K$ to be $70$, which results in an average of $2\rm{,}500$ object proposals on each image. Individually validating and classifying each of them is computationally expensive, so we design a two-step discrimination process, with the first light-weighted classifier identifying most proposals to be `non-object', and the second classifier spending more resource on each survived object and assigning a class label as well as a confidence score to it.

The \textbf{first step} involves training a binary classifier to determine whether each proposal is an object. To this end, we first adopt RoIAlign~\cite{he2017mask} with a kernel size of $7\times7$ to extract the features for each proposal on the box feature map (see Fig.~\ref{fig:pipeline}). Then a $32\times7\times7$ convolution layer is followed to obtain the classification score for each proposal. A binary classifier is built, with the loss function being:
\begin{align}
{\mathcal{L}_\mathrm{prop}}={-\frac{1}{N}\sum_{m=1}^M
\left\{
\begin{aligned}
&\left(1-p_m\right)^\alpha\log\left(p_m\right), \quad\mathrm{if}\ {\mathrm{IoU}_m}\geqslant{\tau}\\
&p_m^\alpha\log\left(1-p_m\right), \quad\quad\ \mathrm{otherwise}\\
\end{aligned}
\right.},
\label{eqn:loss_proposal} 
\end{align}
where $N$ denotes the number of positive samples, $p_m$ denotes the objectness score for the $m$-th proposal, ${p_m}\in{\left[0,1\right]}$, and $\mathrm{IoU}_m$ denotes the maximum IoU value between the $m$-th proposal and all the ground-truth bounding-boxes. $\tau$ is the IoU threshold, set to be $0.7$ throughout this paper. This is to sample a few positive examples to avoid training data imbalance~\cite{lin2017focal}. ${\alpha}={2}$ is a hyper-parameter that smoothes the loss function. According to~\cite{lin2017focal}, we use $\pi=0.1$, so the value of biases is set to $-2.19$.

The \textbf{second step} follows to assign a class label for each survived proposal. This step is very important, since the class labels associated to the corner keypoints are not always reliable. Although, we rely on the corner classes to reject invalid corner pairs, the consensus between them may be incorrect due to the lack of information from the ROI region, so we need a more powerful classifier that incorporates the ROI features to make the final decision.
To this end, we train another classifier with $C$ outputs where $C$ is the number of classes in the dataset. This classifier is also built upon the RoIAlign-features extracted in the first step, but instead extract the features from the category feature map (see Fig.~\ref{fig:pipeline}) to preserve more information and a $C$ dimensional vector is obtained using a $256\times7\times7$ convolution layer, for each of the survived proposals. Then $C$-way classifier is built. A similar loss function considering the class label is used:
\begin{align}
{\mathcal{L}_\mathrm{class}}={-\frac{1}{\hat{N}}\sum_{m=1}^{\hat{M}}\sum_{c=1}^C
\left\{
\begin{aligned}
&\left(1-q_{m,c}\right)^\beta\log\left(q_{m,c}\right), \quad\mathrm{if}\ {\mathrm{IoU}_{m,c}}\geqslant{\tau}\\
&q_{m,c}^\beta\log\left(1-q_{m,c}\right), \quad\quad\ \mathrm{otherwise}\\
\end{aligned}
\right.},
\label{eqn:loss_proposal} 
\end{align}
where $\hat{M}$ and $\hat{N}$ denote the number of survived proposals and the number of positive samples within them, respectively. $\mathrm{IoU}_{m,c}$ denotes the maximum IoU value between the $m$-th proposal and all the ground-truth bounding-boxes of the $c$-th class, and the IoU threshold, $\tau$, remains unchanged. $q_{m,c}$ is the classification score for the $c$-th class of the $m$-th object, and $\beta$ plays a similar role as $\alpha$, and we also fix it to be $2$ in this paper.

Here we emphasize the differences between DeNet~\cite{tychsen2017denet} and our method, although they are similar in the idea level. First, we equip each corner with a multi-class label rather than a binary label, thus we can rely on the class labels to reject the unnecessary invalid corner pairs to save the computational costs of the overall framework. Second, we use an extra lightweight binary classification network to first reduce the number of proposals to be processed by the classification network, while DeNet only relies on one classification network. This helps our method be more efficient. Finally, we design a novel variant of the focal loss for the two classifiers, which is different from the maximum likelihood function in DeNet. This is mainly to solve the significant imbalance between the positive and negative proposals during the training process.

\subsection{The Inference Process}
\label{Approach:Classification}

The inference process simply repeats the training process but uses thresholds to filter out clearly low-quality proposals. Note that even with augmented positive training data, the predicted scores, $p_m$ and $q_{m,c}$, are biased towards $0$. So, in the inference stage, we use a relatively low threshold ($0.2$ in this paper) in the first step to allow more proposals to survive. For each proposal, provided the RoIAlign-features, the computational cost of the first classifier is about $10\%$ of that of the second one. Under the threshold of $0.2$, the average fraction of survived proposals is around $20\%$, making the overheads of these two stages comparable.

For each proposal survived to the second step, we assign it with up to $2$ class labels, corresponding to the dominant class of the corner keypoints and that of the proposal (two classes may be identical, if not, the proposal becomes two proposals with potentially different scores). For each candidate class, we denote $s_1$ 
as  the corner classification score (the average of two corner keypoints, in the range of $\left(0,1\right)$), and $s_2$ as the regional classification score (the probability of the proposal class label, predicted by the multi-class classifier, also in the range of $\left(0,1\right)$). We assume that both scores contribute to the final score, and a positive evidence should be added if either score is larger than $0.5$. Therefore, we compute the score by ${s_c}={\left(s_1+0.5\right)\left(s_2+0.5\right)}$, then we will apply normalization to rescale this score into the $\left[0, 1\right]$. We finally preserve $100$ proposals with highest scores into evaluation. In Table~\ref{tab:cpn_ablation}, we will show that two classifiers provide complementary information and boost the detection accuracy.
\section{Experiments}
\label{Experiments}
\subsection{Dataset and Evaluation Metrics}
\label{Experiments:Dataset}
We evaluate our approach on the MS-COCO dataset~\cite{lin2014microsoft}, one of the most popular object detection benchmarks. It contains a total of $120\mathrm{K}$ images with more than $1.5$ million bounding boxes covering $80$ object categories, making it a very challenging dataset. Following the common practice~\cite{lin2017focal,lin2017feature}, we train our model using the `trainval35k', which is the union set of $80\mathrm{K}$ training images and $35\mathrm{K}$ (a subset of) validation images. We report evaluation results on the standard test-dev set, which has no public annotations, by uploading the results to the evaluation server. For the ablation studies and visualization experiments, we use the $5\mathrm{K}$ validation images remained in the validation set.

We use the average precision (AP) metric defined in MS-COCO to characterize the performance of our approach as well as other competitors. AP computes the average precision over ten IoU thresholds (\textit{i.e.}, $0.5:0.05:0.95$), for all categories. Meanwhile, we follow the convention to report some other important metrics, \textit{e.g.}, $\mathrm{AP_{50}}$ and $\mathrm{AP_{75}}$ are computed at single IoU thresholds of $0.50$ and $0.75$~\cite{everingham2010pascal}, and $\mathrm{AP_{small}}$, $\mathrm{AP_{medium}}$, and $\mathrm{AP_{large}}$ are computed under different object scales (\textit{i.e.}, small for $\mathrm{area}<32^{2}$, medium for $32^{2}<\mathrm{area}<96^{2}$, and large for $\mathrm{area}>96^{2}$), respectively. All metrics are computed by allowing at most $100$ proposals preserved on each testing image.
\vspace{-1ex}
\subsection{Implementation Details}
\label{Experiments:Details}
We implement our method using Pytorch~\cite{paszke2017automatic}, and refer to some codes from CornerNet~\cite{law2018cornernet}, mmdetection~\cite{mmdetection} and CenterNet~\cite{zhou2019objects}. We use CornerNet~\cite{law2018cornernet} and CenterNet~\cite{duan2019centernet} as our baselines. The stacked Hourglass networks~\cite{newell2016stacked} with $52$ and $104$ layers are trained for keypoint extraction, with all modifications made by CornerNet preserved. In addition, we experiment another backbone named DLA-34 ~\cite{yu2018deep}. We follow the modifications made by CenterNet~\cite{zhou2019objects}, but replace the deformable convolutional layers with normal layers.

\noindent\textbf{Training.} All networks are trained from scratch, except the DLA-34, which is initialized with ImageNet pretrain. Cascade corner pooling~\cite{duan2019centernet} is used to help the network better detect corners. The input image is resized into $511\times511$, and the output resolutions for the four feature maps (the top-left and bottom-right heatmaps, the proposal and class feature maps) are all $128\times128$. The data augmentation strategy presented in~\cite{law2018cornernet} is used. The overall loss function is
\begin{align} 
{\mathcal{L}}={\mathcal{L}_\mathrm{det}^\mathrm{corner}+\mathcal{L}_\mathrm{offset}^\mathrm{corner}+\mathcal{L}_\mathrm{prop}+\mathcal{L}_\mathrm{class}},
\label{loss} 
\end{align}
which we use an Adam~\cite{kingma2014adam} optimizer to train our model. On eight NVIDIA Tesla-V100 (32GB) GPUs, we use a batch size of $48$ ($6$ samples on each card) and train the model for $200\mathrm{K}$ iterations with a base learning rate of $2.5\times10^{-4}$ followed by another $50\mathrm{K}$ iterations with a reduced learning rate of $2.5\times10^{-5}$. The training lasts about 9 days, 5 days and 3 days for Hourglass-104, Hourglass-52 and DLA-34, respectively.

\noindent\textbf{Inference.} Following~\cite{law2018cornernet}, both single-scale and multi-scale detection processes are performed. For single-scale testing, we feed the image with the original resolution into the network, while for multi-scale testing, the image is resized into different resolutions ($0.6\times$, $1\times$, $1.2\times$, $1.5\times$, and $1.8\times$) and then fed into the network. Flip argumentation is added to both single-scale or multi-scale evaluation. For the multi-scale evaluation, the predictions for all scales (including the flipped variants) are fused into the final result. we use soft-NMS~\cite{bodla2017soft} to suppress the redundant bounding-boxes, and preserve $100$ top-scored bounding-boxes for final evaluation.
\vspace{-2ex}
\subsection{Comparisons with State-of-the-Art Detectors}
\label{Experiments:Comparisons}

\begin{table}[!t]
\small
\begin{center}
\caption{Inference accuracy ($\%$) of CPN and state-of-the-art detectors on the COCO \textit{test-dev} set. CPN ranks among the top of state-of-the-art detectors. `R', `X', `HG', `DCN' and `$\dagger$' denote ResNet, ResNeXt, Hourglass, Deformable Convolution Network~\cite{dai2017deformable}, and multi-scale training or testing, respectively.}
\vspace{-3ex}
\label{tab:comparision}
\resizebox{1.0\textwidth}{!}{
\begin{tabular}{|l|c|c|cccccc|}
\hline
Method & Backbone & Input Size & AP & AP$_{50}$ & AP$_{75}$ & AP$_\mathrm{S}$ & AP$_\mathrm{M}$ & AP$_\mathrm{L}$\\
\hline
\hline
\textbf{Anchor-based:} & & & & & & & &\\
Faster R-CNN~\cite{lin2017feature} & R-101 & 600 & 36.2 & 59.1 & 39.0 & 18.2 & 39.0 & 48.2\\
RetinaNet~\cite{lin2017focal} & R-101 & 800 & 39.1 & 59.1 & 42.3 & 21.8 & 42.7 & 50.2\\       
Cascade R-CNN~\cite{cai2018cascade} & R-101 & 800 & 42.8 & 62.1 & 46.3 & 23.7 & 45.5 & 55.2\\
Libra R-CNN~\cite{pang2019libra} & X-101-64x4d & 800 & 43.0 & 64.0 & 47.0 & 25.3 & 45.6 & 54.6\\
Grid R-CNN~\cite{lu2019grid} & X-101-64x4d & 800 & 43.2 & 63.0 & 46.6 & 25.1 & 46.5 & 55.2\\
YOLOv4~\cite{bochkovskiy2020yolov4} & CSPDarknet-53 & 608 & 43.5 & 65.7 & 47.3 & 26.7 & 46.7 & 53.3\\
AlignDet~\cite{chen2019revisiting} & X-101-32x8d & 800 & 44.1 & 64.7 & 48.9 & 26.9 & 47.0 & 54.7\\
AB+FSAF~\cite{zhu2019feature}~$\dagger$ & X-101-64x4d & 800 & 44.6 & 65.2 & 48.6 & 29.7 & 47.1 & 54.6\\
FreeAnchor~\cite{zhang2019freeanchor}~$\dagger$ & X-101-32x8d & $\le$1280 & 47.3 & 66.3 & 51.5 & 30.6 & 50.4 & 59.0\\
PANet~\cite{liu2018path}~$\dagger$ & X-101-64x4d & 840 & 47.4 & 67.2 & 51.8 & 30.1 & 51.7 & 60.0\\
TridentNet~\cite{li2019scale}~$\dagger$ & R-101-DCN & 800 & 48.4 & 69.7 & 53.5 & 31.8 & 51.3 & 60.3\\
ATSS~\cite{zhang2020bridging}~$\dagger$ & X-101-64x4d-DCN & 800 & 50.7 & 68.9 & 56.3 & 33.2 & 52.9 & 62.4\\
EfficientDet~\cite{tan2020efficientdet}& EfficientNet~\cite{tan2019efficientnet} & 1536 & ~\textbf{53.7} & ~\textbf{72.4} & ~\textbf{58.4} & - & - & -\\    
\hline
\hline
\textbf{Anchor-free:} & & & & & & & &\\
GA-Faster-RCNN~\cite{wang2019region} & R-50 & 800 & 39.8 & 59.2 & 43.5 & 21.8 & 42.6 & 50.7\\
FoveaBox~\cite{kong2020foveabox} & R-101 & 800 & 42.1 & 61.9 & 45.2 & 24.9 & 46.8 & 55.6\\
ExtremeNet~\cite{zhou2019bottom}~$\dagger$ & HG-104 & $\le$1.5$\times$ & 43.2 & 59.8 & 46.4 & 24.1 & 46.0 & 57.1\\
FCOS~\cite{tian2019fcos} w/ imprv & X-101-64x4d & 800 & 44.7 & 64.1 & 48.4 & 27.6 & 47.5 & 55.6\\
CenterNet~\cite{zhou2019objects}~$\dagger$ & HG-104 & $\le$1.5$\times$ & 45.1 & 63.9 & 49.3 & 26.6 & 47.1 & 57.7\\
RPDet~\cite{yang2019reppoints}~$\dagger$ & R-101-DCN & 800 & 46.5 & 67.4 & 50.9 & 30.3 & 49.7 & 57.1\\
SAPD~\cite{zhu2019soft} & X-101-64x4d & 800 & 45.4 & 65.6 & 48.9 & 27.3 & 48.7 & 56.8\\
SAPD~\cite{zhu2019soft} & X-101-64x4d-DCN & 800 & 47.4 & 67.4 & 51.1 & 28.1 & 50.3 & 61.5\\
\hline
CornerNet~\cite{law2018cornernet} & HG-104 & ori. & 40.5 & 56.5 & 43.1 & 19.4 & 42.7 & 53.9\\
CornerNet~\cite{law2018cornernet}~$\dagger$ & HG-104 & $\le$1.5$\times$ & 42.1 & 57.8 & 45.3 & 20.8 & 44.8 & 56.7\\
CenterNet~\cite{duan2019centernet}& HG-104 & ori. & 44.9 & 62.4 & 48.1 & 25.6 & 47.4 & 57.4\\
CenterNet~\cite{duan2019centernet}~$\dagger$ & HG-104 & $\le$1.8$\times$ & 47.0 & 64.5 & 50.7 & 28.9 & 49.9 & 58.9\\
\hline
\rowcolor{black!20}
\textbf{CPN} & DLA-34 & ori. & 41.7 & 58.9 & 44.9 & 20.2 & 44.1 & 56.4\\
\rowcolor{black!20}
\textbf{CPN} & HG-52 & ori. & 43.9 & 61.6 & 47.5 & 23.9 & 46.3 & 57.1\\
\rowcolor{black!20}
\textbf{CPN} & HG-104 & ori. & 47.0 & 65.0 & 51.0 & 26.5 & 50.2 & 60.7\\
\rowcolor{black!20}
\textbf{CPN}~$\dagger$ & DLA-34 & $\le$1.8$\times$ & 44.5 & 62.3 & 48.3 & 25.2 & 46.7 & 58.2\\
\rowcolor{black!20}
\textbf{CPN}~$\dagger$ & HG-52 & $\le$1.8$\times$ & 45.8 & 63.9 & 49.7 & 26.8 & 48.4 & 59.4\\
\rowcolor{black!20}
\textbf{CPN}~$\dagger$ & HG-104 & $\le$1.8$\times$ & \textbf{49.2} & \textbf{67.4} & \textbf{53.7} & \textbf{31.0} & \textbf{51.9} & \textbf{62.4}\\
\hline
\end{tabular}}
\vspace{-4ex}
\end{center}
\end{table}

We report the inference accuracy of CPN on the MS-COCO test-dev set and compare with the state-of-the-art detectors, as shown in Table~\ref{tab:comparision}. CPN obtains a significant improvement compared to CornerNet~\cite{law2018cornernet} and CenterNet~\cite{duan2019centernet}, two direct baselines. Specifically, CPN-52 (indicating that the backbone is Hourglass-52) reports a single-scale testing AP of $43.9\%$ and a multi-scale testing AP of $45.8\%$, which surpasses CornerNet-104, with a deeper backbone, by $3.4\%$ and $3.7\%$, respectively. Equipped with a deeper backbone (\textit{i.e.}, Hourglass-104), CPN demonstrate a large advantage over CenterNet, the previous best anchor-free detector, with margins of $2.1\%$ and $2.2\%$ under single-scale and multi-scale settings, respectively.

CPN also takes the lead in other metrics. For example, AP$_{50}$ and AP$_{75}$ reflect the accuracy of proposal localization and class prediction. Compared to CenterNet, CPN reports higher AP scores especially for AP$_{75}$ (\textit{e.g.}, CPN-104 reports a single-scale testing AP$_{75}$ of $51.0\%$, claiming an improvement of $2.9\%$ over CenterNet). This suggests that some inaccurate bounding boxes with IoU value between $0.5$ and $0.7$ are difficult for CenterNet to filter out with merely center information incorporated. AP$_\mathrm{S}$, AP$_\mathrm{M}$ and AP$_\mathrm{L}$ reflect the detection accuracy for objects with different scales. CPN improves more for AP$_\mathrm{M}$ and AP$_\mathrm{L}$ than AP$_\mathrm{S}$ (\textit{e.g.}, CPN-104 reports single-scale testing AP$_\mathrm{S}$, AP$_\mathrm{M}$ and AP$_\mathrm{L}$ of $26.5\%$, $50.2\%$ and $60.7\%$, which improves by $0.9\%$, $2.8\%$ and $3.3\%$ from CenterNet, respectively). This is because medium and large objects require richer semantic information to be extracted from the proposal, which is not likely to be handled well with a center keypoint.

When comparing with other anchor-free approaches, CPN-52 reports a single-scale testing AP of $43.9\%$, which is already better than some of anchor-free detector with deeper backbones (\textit{e.g.}, FoveaBox~\cite{kong2020foveabox} and ExtremeNet~\cite{zhou2019bottom}~$\dagger$). The best performance of CPN reaches an AP of $49.2\%$, surpassing all published anchor-free approaches to the best of our knowledge. Meanwhile, CPN is also competitive among anchor-based detectors, \textit{e.g.}, CPN-52 reports a single-scale testing AP of $43.9$\% which is comparable to $44.1\%$ of AlignDet~\cite{chen2019revisiting}, and CPN-104 reports a single-scale testing AP of $47.0\%$ which is comparable $47.4\%$ of PANet~\cite{liu2018path}. FreeAnchor~\cite{zhang2019freeanchor} used a larger resolution of input images (\textit{e.g.}, $\sim800\times1300$) to improve the AP of small objects, while CPN-104 with multi-scale testing outperforms it on AP$_\mathrm{S}$ by a margin of $0.4\%$.

\subsection{Classification Improves Precision}
\label{Experiments:Classifiers}

\begin{table}[!t]
\small
\centering
\caption{The detection performance ($\%$) of different classification options on CPN.}
\label{tab:cpn_ablation}
\renewcommand\tabcolsep{0.1cm} 
\begin{tabular}{|c|cc|cccccc|c|}
\hline
Backbone & B-Classifier & M-Classifier & AP & AP$_{50}$ & AP$_{75}$ & AP$_\mathrm{S}$ & AP$_\mathrm{M}$ &  AP$_\mathrm{L}$ & AR$_\mathrm{100}$\\
\hline
\hline
\multirow{4}{*}{HG-52} &             &            & 38.6   & 54.5  & 41.4 & 19.0  & 40.6 & 54.4 & 57.4\\
                     & \checkmark &            & 42.0   & 59.7  & 45.1 &  24.2 & 45.0 & 56.6 & 60.5\\
                     &            & \checkmark & 42.4 & 60.2 & 45.7 & 23.3 & 44.7 & 58.6 & 59.0\\
                     & \checkmark & \checkmark & \textbf{43.8} & \textbf{61.4} & \textbf{47.4} & 
                                                 \textbf{24.7} & \textbf{46.5} & \textbf{60.0} & 
                                                 \textbf{60.9}\\
\hline
\end{tabular}
\vspace{-1ex}
\end{table}

We investigate the improvement of precision brought by the classification stage. Note that there are two classifiers, with the first one (binary) determines the objectness of each proposal, and the second one (multi-class) providing complementary information of the class label. We perform ablation study in Table~\ref{tab:cpn_ablation} to analyze the contribution of individual classifiers -- in the scenarios that the multi-class classifier is missing, we directly use the class label of the corner keypoints as the final prediction. On the one hand, both classifiers can improve the AP of the basic model (one-stage corner keypoint grouping) significantly ($3.4\%$ and $3.8\%$ absolute gains). On the other hand, two classifiers provide complementary information, so that combining them can further improve the AP by more than $1\%$. This indicates that the semantic information required for determining the objectness and class label of a proposal is slightly different.

\begin{table}[!t]
\small
\centering
\vspace{-1ex}
\caption{We report the average false discovery rates ($\%$, lower is better) for CornerNet, CenterNet and CPN on the MS-COCO validation dataset. The results show that our approach generates fewer false positives. Under the same corner keypoint extractor, this is the key to outperform the baselines in the AP metrics.}
\label{tab:AF}
\renewcommand\tabcolsep{0.08cm} 
\begin{tabular}{|l|c|cccc|ccc|}
\hline
Method & Backbone & AF & AF$_{5}$ & AF$_{25}$ & AF$_{50}$ & AF$_\mathrm{S}$ & AF$_\mathrm{M}$ & AF$_\mathrm{L}$\\
\hline
\hline
CornerNet~\cite{law2018cornernet} & HG-52 & 40.4 & 35.2 & 39.4 & 46.7 & 62.5 & 36.9 & 28.0\\
CenterNet~\cite{duan2019centernet} & HG-52 & 35.1 & 30.7 & 34.2 & 40.8 & 53.0 & 31.3 & 24.4\\
CPN                        & HG-52 & \textbf{33.4} & \textbf{29.5} & \textbf{32.5} & \textbf{38.6} & 
                                     \textbf{52.0} & \textbf{29.2} & \textbf{21.0}\\
\hline
CornerNet~\cite{law2018cornernet} & HG-104 & 37.8 & 32.7 & 36.8 & 43.8 & 60.3 & 33.2 & 25.1\\
CenterNet~\cite{duan2019centernet} & HG-104& 32.4 & 28.2 & 31.6 & 37.5 & 50.7 & 27.1 & 23.0\\
CPN                        & HG-104 & \textbf{30.6} & \textbf{26.9} & \textbf{29.7} & \textbf{35.5} & 
                                      \textbf{48.8} & \textbf{25.7} & \textbf{19.2}\\
\hline
\end{tabular}
\vspace{-1ex}
\end{table}

We have discussed the importance of the incorrect bounding box suppression for the improvement of detection accuracy and recall in Section~\ref{Approach:Choices}. For a more intuitive analysis of false positives, we adopt the average false discovery rate (AF) metric\footnote{$\mathrm{AF}=1-\mathrm{\widetilde{AP}}$, where $\mathrm{\widetilde{AP}}$ is computed over IoU thresholds of $\left[0.05:0.05:0.5\right]$ for all categories. ${\mathrm{AF}_{\tau}}={1-\mathrm{\widetilde{AP}}_{\tau}}$, where $\mathrm{\widetilde{AP}}_{\tau}$ is computed at the IoU threshold of $\tau\%$, ${\mathrm{AF_{scale}}}={1-\mathrm{\widetilde{AP}_{scale}}}$, where ${\mathrm{scale}}={\left\{\mathrm{small},\mathrm{medium},\mathrm{large}\right\}}$, indicating the scale of object.}~\cite{duan2019centernet} to quantify the fraction of incorrectly grouped proposals for different detectors. Results are shown in Table~\ref{tab:AF}. CPN-52 and CPN-104 report AF of $33.4\%$ and $30.6\%$, respectively, which are lower than the direct baselines, CornerNet and CenterNet.

\begin{table}[!h]
	\small
	\small
	\centering
	\caption{The detection performance ($\%$) of using different ways (instance embedding and binary classification) to determine the validity of a proposal.}
	\label{tab:classifier_embedding}
	\renewcommand\tabcolsep{0.08cm} 
	\begin{tabular}{|c|c|c|cccccc|c|}
		\hline
		Method & Backbone & Objectness & AP & AP$_{50}$ & AP$_{75}$ & AP$_\mathrm{S}$ & AP$_\mathrm{M}$ &  AP$_\mathrm{L}$ & AR$_\mathrm{100}$\\
		\hline
		\hline
		\multirow{2}*{CornerNet~\cite{law2018cornernet}} & \multirow{2}*{HG-52} & Embedding & 38.3  & 54.2  & 40.5 & 18.5  & 39.6 & 52.2 & 56.7\\
		&                  & B-Classifier & \textbf{42.3} & \textbf{59.5}  &
		\textbf{45.4} & \textbf{24.6} & \textbf{45.4} & \textbf{57.6} & \textbf{59.9}\\
		\hline                                    
		\multirow{2}*{CenterNet~\cite{duan2019centernet}} & \multirow{2}*{HG-52} & Embedding & 41.3 & 59.2 & 43.9 & 23.6 & 43.6 & 53.6 & 59.0\\
		&                      & B-Classifier & \textbf{42.6} & \textbf{59.8} & \textbf{45.8} & \textbf{25.1} & \textbf{45.7} & \textbf{57.7} & \textbf{60.1}\\
		\hline
	\end{tabular}
\end{table}

Note that these three methods share a similar way of extracting corner keypoints, but CornerNet suffers large AF values due to the lack of validation beyond the proposals. CenterNet, by forcing a center keypoint to be detected, was believed effective in filtering out false positives, and our approach, by reevaluating the proposal based on regional features, performs better than CenterNet. More importantly, by inserting an individual classification stage, CPN alleviates the false positives caused by the instance embedding mechanism, as shown in Table~\ref{tab:classifier_embedding}.

\begin{figure}[!t]
\centering 
\includegraphics[width=1\textwidth]{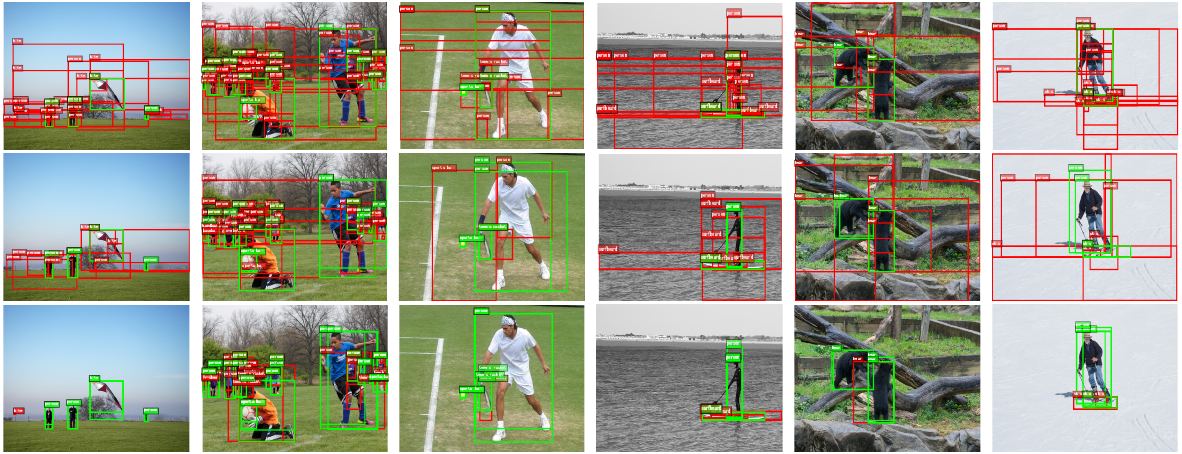}
\caption{Typical detection results showing how CPN filters out false positives detected by CornerNet~\cite{law2018cornernet} and CenterNet~\cite{duan2019centernet}. In each group, from top to bottom are the detection results of CornerNet, CenterNet and CPN, respectively. The green and red bounding-boxes denote true positives and false positives, respectively.}
\label{fig:ablation_comparisons} 
\vspace{-1ex}
\end{figure}

Last but not least, we provide some visualization results in Figure~\ref{fig:ablation_comparisons} to show that CPN indeed enjoys a strong ability in improving the precision of detection. We compare against three keypoint-grouping-based detectors. CornerNet, merely relying on instance embedding, found many seemingly strange `objects'. CenterNet, by enforcing a center keypoint to be found in the central region, eliminated a part of them but some of the false positives can still survive. Using an individual classifier, as CPN does, is more powerful in filtering out false positives. Yet, as we shall see in the next part, CPN enjoys faster inference speed than both CornerNet and CenterNet by efficiently sharing computation between two stages.

\subsection{Inference Speed}
\label{Experiments:Speed}

To show that CPN can generate high-quality bounding boxes with small computational costs, we report the inference speed for CPN on the MS-COCO validation dataset under different settings and compare the results to state-of-the-art efficient detectors, as shown in Table~\ref{tab:speed}. For fair comparison, we test the inference speed for all detectors on an NVIDIA Tesla-V100 GPU. CPN-104 reports an FPS/AP of $7.3/46.8\%$, which is both faster and better than CenterNet-104 ($5.1/44.8\%$) under the same setting. With a lighter backbone of Hourglass-52, CPN-52 reports an FPS/AP of $9.9/43.8\%$, which outperforms both CornerNet-52 and CenterNet-52. This indicates that two-stage detectors are not necessarily slow -- our solution, by sharing a large amount of computation between the first (for keypoint extraction) and the second (for feature extraction) stage, achieves a good trade-off between inference speed and accuracy.

In the scenarios that require faster inference speed, CPN can be further accelerated by replacing with a lighter backbone and not using flip augmentation at the inference stage. In this configuration, CPN-34 (indicating that the backbone is DLA-34) reports FPS/AP of $43.3/39.7\%$ and $26.2/41.6\%$, respectively, surpassing other competitors with similar computational complexity.

\begin{table*}[!t]
\footnotesize
\centering
\caption{Inference speed of CPN under different conditions \textit{vs.} other detectors on the MS-COCO validation dataset. FPS is measured on the on an NVIDIA Tesla-V100 GPU. CPN achieves a good trade-off between accuracy and speed.}
\label{tab:speed}
\renewcommand\tabcolsep{0.15cm} 
\begin{tabular}{|l|c|c|c|c|c|}
\hline
Method & Backbone & Input Size & Flip & AP & FPS\\
\hline
FCOS~\cite{tian2019fcos} & X-101-64x4d & 800 & $\times$ & 42.6 & 8.1\\
Faster R-CNN~\cite{ren2015faster} & X-101-64x4d & 800 & $\times$ & 41.1 & 8.2\\
CornerNet-lite~\cite{law2019cornernet} & HG-Squeeze & ori. & \checkmark & 36.5 & 22.0\\
\hline
CornerNet~\cite{law2018cornernet} & HG-104 & ori. & \checkmark & 41.0 & 5.8\\
CenterNet~\cite{duan2019centernet} & HG-104 & ori. & \checkmark & 44.8 & 5.1\\
\hline
\hline
CPN & HG-104 & ori. & \checkmark & 46.8 & 7.3\\
CPN & HG-52 & ori. & \checkmark & 43.8 & 9.9\\
CPN & HG-104 & $0.7\times$ori. & $\times$ & 40.5 & 17.9\\
CPN & DLA-34 & ori. & \checkmark & 41.6 & 26.2\\
CPN & DLA-34 & ori. & $\times$ & 39.7 & 43.3\\
\hline
\end{tabular}
\end{table*}

\section{Conclusions}

In this paper, we present an anchor-free, two-stage object detection framework. It starts with extracting corner keypoints and composing them into object proposals, and applies two-step classification to filter out false positives. With the above two stages, the recall and precision of detection are significantly improved, and the final result ranks among the top of existing object detection methods.

The most important take-away is that anchor-free methods are more flexible in proposal extraction, while an individual discrimination stage is required to improve precision. When implemented properly, such a two-stage framework can be efficient in evaluation. Therefore, the debate on using one-stage or two-stage detectors seems not critical.

%
%
\bibliographystyle{splncs04}
\bibliography{egbib}
\end{document}